\documentclass[letterpaper]{article} 
\usepackage{aaai19}  
\usepackage{times}  
\usepackage{helvet}  
\usepackage{courier}  
\usepackage{url}  
\usepackage{graphicx}  
\usepackage{latexsym}
\usepackage{amsmath}
\usepackage{amsfonts}
\usepackage{algpseudocode, algorithmicx, algorithm}
\usepackage{array}
\usepackage{booktabs}
\usepackage{bigstrut}
\newcolumntype{L}[1]{>{\raggedright\let\newline\\\arraybackslash\hspace{0pt}}m{#1}}
\newcolumntype{C}[1]{>{\centering\let\newline\\\arraybackslash\hspace{0pt}}m{#1}}
\newcolumntype{R}[1]{>{\raggedleft\let\newline\\\arraybackslash\hspace{0pt}}m{#1}}

\title{Learning to Embed Sentences Using Attentive Recursive Trees}
\author{Jiaxin Shi,\textsuperscript{1}
Lei Hou,\textsuperscript{1}\protect \thanks{Corresponding author.}
Juanzi Li,\textsuperscript{1}
Zhiyuan Liu,\textsuperscript{1}
Hanwang Zhang\textsuperscript{2}\\
\textsuperscript{1}{Tsinghua University}\\
\textsuperscript{2}{Nanyang Technological University}\\
shijx12@gmail.com,
\{houlei,lijuanzi,liuzy\}@tsinghua.edu.cn,
hanwangzhang@ntu.edu.sg
}

\begin{document}
\maketitle
\begin{abstract}

Sentence embedding is an effective feature representation for most deep learning-based NLP tasks. One prevailing line of methods is using recursive latent tree-structured networks to embed sentences with task-specific structures. However, existing models have no explicit mechanism to emphasize task-informative words in the tree structure. To this end, we propose an Attentive Recursive Tree model (AR-Tree), where the words are dynamically located according to their importance in the task. Specifically, we construct the latent tree for a sentence in a proposed important-first strategy, and place more attentive words nearer to the root; thus, AR-Tree can inherently emphasize important words during the bottom-up composition of the sentence embedding. We propose an end-to-end reinforced training strategy for AR-Tree, which is demonstrated to consistently outperform, or be at least comparable to, the state-of-the-art sentence embedding methods on three sentence understanding tasks.
\end{abstract}

\section{Introduction}

Along with the success of representation learning (\textit{e.g.}, word2vec~\cite{mikolov2013distributed}), sentence embedding, which maps sentences into dense real-valued vectors that represent their semantics, has received much attention. It is playing a critical role in many applications such as sentiment analysis~\cite{socher2013recursive}, question answering~\cite{wang2015long} and entailment recognition~\cite{bowman2015large}.

\begin{figure}[ht]
\includegraphics[width=\linewidth]{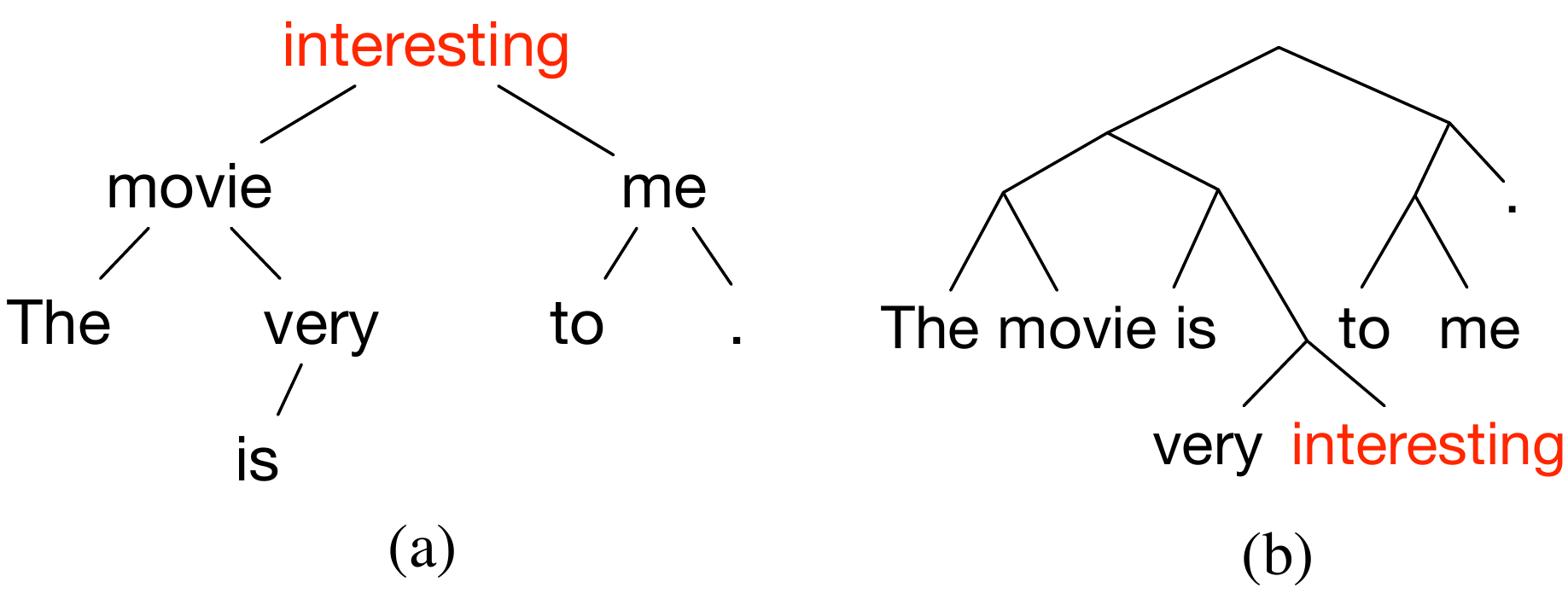}
\caption{Two recursive trees for sentence \emph{The movie is very interesting to me.} in the sentiment analysis task. Our AR-Tree (a) is constructed by recursively selecting the most informative word, \textit{e.g.}, \emph{interesting}. However, other latent trees (b) are built by composing adjacent pairs, \textit{e.g.}, \emph{very interesting} captured by \cite{choi2017learning}, which lacks the potential to emphasize words.}
\label{fig:1}
\end{figure}

There are three predominant approaches for constructing sentence embeddings. (1) Recurrent neural networks (RNNs) encode sentences word by word in sequential order~\cite{dai2015semi,hill2016learning}. (2) Convolutional neural networks (CNNs) produce sentence embeddings in a bottom-up manner, moving from local n-grams to the global sentence as the receptive fields enlarge~\cite{blunsom2014convolutional,hu2014convolutional}. However, the above two approaches cannot well encode linguistic composition of natural languages to some extent. (3) The last approach, on which this paper focuses, exploits tree-structured recursive neural networks (TreeRNNs)~\cite{socher2011semi,socher2013recursive} to embed a sentence along its parsing tree. Tree-structured Long Short-Term Memory (Tree-LSTM)~\cite{tai2015improved,zhu2015long} is one of the most renowned variants of TreeRNNs that is shown to be effective in learning task-specific sentence embeddings~\cite{bowman2016fast}.


Tree-LSTM models are motivated by the intuition that in human languages there are complicated hierarchical structures which contain rich semantics. 
Latent tree models~\cite{yogatama2016learning,maillard2017jointly,choi2017learning,williams2017learning} can learn the optimal hierarchical structure, which may vary from tasks to tasks, without explicit structure annotations.
The training signals to parse and embed sentences are both from certain downstream tasks.
Existing models place all words in leaves equally and build the tree structure and the sentence embedding by composing adjacent node pairs bottom up (\textit{e.g.}, Figure~\ref{fig:1}b). 
This mechanism prevents the sentence embedding from focusing on the most informative words, resulting in a performance limitation on certain tasks~\cite{shi2018tree}.


To address this issue, we propose an Attentive Recursive Tree model (\textbf{AR-Tree}) for sentence embedding, which is a novel framework that incorporates task-specific attention mechanism into the latent tree structure learning~\cite{dos2016attentive}. 
AR-Tree represents a sentence as a binary tree that contains one word in each leaf and non-leaf node, similar to the dependency parsing tree~\cite{nivre2003dependency} but our AR-Tree does not depend on manual rules.
To utilize the sequential information, we expect the tree's in-order traversal preserves the word sequence, so that we can easily recover the original word sequence and obtain context of a word from its subtrees.
As shown in Figure~\ref{fig:1}a, the key advantage of an AR-Tree is that those task-important words will be placed at those nodes near the root and will be naturally emphasized in tree-based embedding. 
This is attributed to our proposed top-down attention-first parsing strategy, inspired by easy-first parsing~\cite{goldberg2010efficient}. 
Specifically, we introduce a trainable scoring function to measure the word attention in a sentence with respect to a task. 
We greedily select the word with the highest score (\textit{e.g.}, \emph{interesting}) as the root node and then recursively parse the remaining two subsequences (\textit{e.g.}, \emph{The movie is} and \emph{to me.}) to obtain two children of the parent node. 
After the tree construction, we embed the sentence using a modified Tree-LSTM unit~\cite{tai2015improved,zhu2015long} in a bottom-up manner, \textit{i.e.}, the resultant embedding is obtained at the root node and is then applied in a downstream application.
As the Tree-LSTM computes node vectors incrementally from leaf nodes to the root node, our model naturally pays more attention to those shallower words, i.e., task-informative words, meanwhile remaining advantages of the recursive semantic composition~\cite{socher2013recursive,zhu2015long}.

Training AR-Tree is challenging due to the non-differentiability caused by the dynamic decision-making procedure.
To this end, we develop a novel end-to-end training strategy based on REINFORCE algorithm~\cite{williams1992simple}.
To make REINFORCE work for the structure inference, we equip it with a weighted reward which is sensitive to the tree structures and a macro normalization strategy for the policy gradients.


We evaluate our model on three benchmarking tasks: textual entailment, sentiment classification, and author profiling. We show that AR-Tree outperforms previous Tree-LSTM models and is comparable to other state-of-the-art sentence embedding models. Further qualitative analyses demonstrate that AR-Tree learns reasonable task-specific attention structures.

To sum up, the contributions of our work are as follows:

\begin{itemize}
    \item We propose Attentive Recursive Tree (AR-Tree), a Tree-LSTM based sentence embedding model, which can parse the latent tree structure dynamically and emphasize informative words inherently.
    \item We design a novel REINFORCE algorithm for the training of discrete tree parsing.
    \item We demonstrate that AR-Tree outperforms previous Tree-LSTM models and is comparable to other state-of-the-art sentence embedding models in three benchmarks.
\end{itemize}

\section{Related Work}

\textbf{Latent Tree-Based Sentence Embedding.}
\cite{bowman2016fast} build trees and compose semantics via a generic shift-reduce parser, whose training relies on ground-truth parsing trees. In this paper, we are interested in latent trees that dynamically parse a sentence without syntax supervision. Combination of latent tree learning with TreeRNNs has been shown as an effective approach for sentence embedding as it jointly optimizes the sentence compositions and a task-specific objective. 
For example, \cite{yogatama2016learning} use reinforcement learning to train a shift-reduce parser without any ground-truth.

\citeauthor{maillard2017jointly} use a CYK chart parser~\cite{cocke1970programming,younger1967recognition,kasami1965efficient} instead of the shift-reduce parser and make it fully differentiable with the help of the softmax annealing technique. However, their model suffers from both time and space issues as the chart parser requires $\mathcal{O}(n^3)$ time and space complexity. \cite{choi2017learning} propose an easy-first parsing strategy, which scores each adjacent node pair using a query vector and greedily combines the best pair into one parent node at each step. They use Straight-Through Gumbel-Softmax estimator~\cite{jang2016categorical} to compute parent embedding in a hard categorical gating way and enable the end-to-end training. \cite{williams2017learning} compare above-mentioned models on several datasets and demonstrate that \cite{choi2017learning} achieve the best performance.

\textbf{Attention-Based Sentence Embedding.} Attention-based methods can be divided into two categories: inter-attention~\cite{dos2016attentive,munkhdalai2017neural}, which requires a pair of sentences to attend with each other, and intra-attention~\cite{arora2016simple,lin2017structured}, which does not require extra inputs except the sentence; thus the latter is more flexible than the former. \cite{kim2017structured} incorporate structural distributions into attention networks using graphical models instead of recursive trees. Note that existing latent tree-based models treat all input words equally as leaf nodes and ignore the fact that different words make varying degrees of contributions to the sentence semantics, which is nevertheless the fundamental motivation of attention mechanism. To our best knowledge, AR-Tree is the first model that generates attentive tree structures and allows the TreeRNNs to focus on more informative words for sentence embeddings.

\section{Attentive Recursive Tree}\label{sec:model}

We represent an input sentence $S$ of $N$ words as $\{\mathbf{x}_1, \mathbf{x}_2, \cdots, \mathbf{x}_N\}$, where $\mathbf{x}_i$ is a $D_x$-dimensional word embedding vector.
For each sentence, we build an \emph{Attentive Recursive Tree} (AR-Tree) where the root and nodes are denoted by $R$ and $T$, respectively.
Each node $t \in T$ contains one word denoted as $t.index$ ($t.index = i$ means the $i$-th word of input sentence) and has two children denoted by $t.left \in T$ and $t.right \in T$ ($nil$ for missing cases). 
Following previous work~\cite{choi2017learning}, we discuss binary trees in this paper and leave the n-ary case for future work.
To keep the important sequential information, we guarantee that the \emph{in-order} traversal of $T$ corresponds to $S$ (i.e., all nodes in $t$'s left subtree must contain an index less than $t.index$).
The most outstanding property of AR-Tree is that words with more task-specific information are closer to the root.

To achieve the property, we devise a scoring function to measure the degree of importance of words, and recursively select the word with the maximum score in a top-down manner. To obtain the sentence embedding, we apply a modified Tree-LSTM to embed the nodes bottom-up, \textit{i.e.}, from leaf to root. The resultant sentence embedding is fed into downstream tasks.

\subsection{Top-Down AR-Tree Construction}
We feed the input sentence into a bidirectional LSTM and obtain a context-aware hidden vector for each word:
\begin{equation}
\begin{aligned}
\overrightarrow{\mathbf{h}_i}, \overrightarrow{\mathbf{c}_i} &= \overrightarrow{\text{LSTM}}(\mathbf{x}_i, \overrightarrow{\mathbf{h}_{i-1}}, \overrightarrow{\mathbf{c}_{i-1}}),\\
\overleftarrow{\mathbf{h}_i}, \overleftarrow{\mathbf{c}_i} &= \overleftarrow{\text{LSTM}}(\mathbf{x}_i, \overleftarrow{\mathbf{h}_{i+1}}, \overleftarrow{\mathbf{c}_{i+1}}),\\
\mathbf{h}_i &= [\overrightarrow{\mathbf{h}_i}; \overleftarrow{\mathbf{h}_i}],\\
\mathbf{c}_i &= [\overrightarrow{\mathbf{c}_i}; \overleftarrow{\mathbf{c}_i}],
\end{aligned}
\end{equation}

where $\mathbf{h}, \mathbf{c}$ denote the hidden states and the cell states respectively. We utilize $\mathbf{h}$ for scoring and let $S=\{\mathbf{h}_1, \mathbf{h}_2, \cdots, \mathbf{h}_N\}$.
Based on these context-aware word embeddings, we design a trainable scoring function to reflect the importance of each word:
\begin{equation}\label{formula:score}
Score(\mathbf{h}_i) = \text{MLP}(\mathbf{h}_i; \theta),
\end{equation}
where MLP can be any multi-layer perceptron parameterized by $\theta$. In particular, we use a 2-layer MLP with 128 hidden units and ReLU activation.
Traditional tf-idf is a simple and intuitive method to reflect the degree of importance of words, however, it is not designed for specific tasks.
We will use it as a baseline.

\begin{algorithm}
\caption{Recursive AR-Tree construction\label{construct}}
\textbf{Input:} Sentence hidden vectors $S=\{\mathbf{h}_1, \mathbf{h}_2, \cdots, \mathbf{h}_N\}$, beginning index $b$ and ending index $e$\\
\textbf{Output:} root node $R_{S[b:e]}$ of sequence $S[b:e]$
\begin{algorithmic}
\Procedure{build}{$S, b, e$}
    \State $R \gets nil$
    \If{$e=b$}
        \State $R \gets$ new Node
        \State $R.index \gets b$
        \State $R.left, R.right \gets nil, nil$
    \ElsIf{$e > b$}
        \State $R \gets$ new Node
        \State $R.index \gets argmax_{i=b}^e Score(\mathbf{h}_i)$
        \State $R.left \gets \Call{build}{S, b, R.index-1}$
        \State $R.right \gets \Call{build}{S, R.index+1, e}$
    \EndIf
    \State \Return $R$
\EndProcedure
\end{algorithmic}
\end{algorithm}

We use a recursive top-down attention-first strategy to construct AR-Tree. Given an input sentence $S$ and the scores for all the words, we select the word with the maximum score as the root $R$ and recursively deal with the remaining two subsequences (before and after the selected word) to obtain its two children. 
Algorithm~\ref{construct} gives the procedure of constructing AR-Tree for sequence $S[b:e] = \{\mathbf{h}_b, \mathbf{h}_{b+1}, \cdots, \mathbf{h}_e\}$. 
We can obtain the whole sentence's AR-Tree by calling $R = \text{BUILD}(S, 1, N)$ and obtain $T$ by the traversal of all nodes. 
In the parsed AR-Tree, each node is most informative among its rooted subtree. Note that we do not use any extra information during the construction, thus AR-Tree is generic for any sentence embedding task.

\subsection{Bottom-Up Tree-LSTM Embedding}\label{sec:bottom-up}
\begin{figure}[ht]
\includegraphics[width=\linewidth]{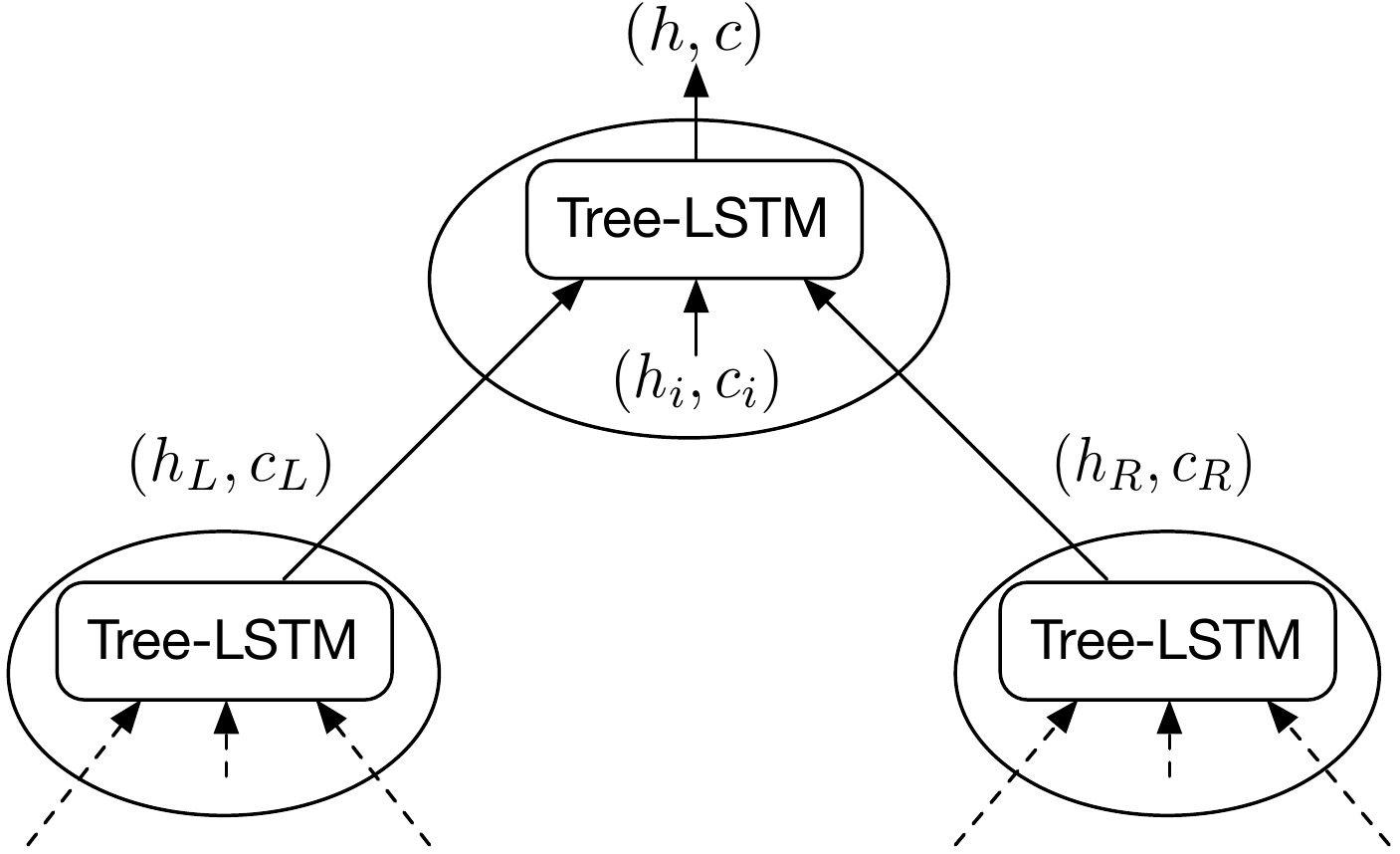}
\caption{Our Tree-LSTM unit composes semantics of left child ($\mathbf{h_L}$,$\mathbf{c_L}$), right child ($\mathbf{h_R}$,$\mathbf{c_R}$) and current word ($\mathbf{h_i}$,$\mathbf{c_i}$) to obtain the node embedding ($\mathbf{h}$,$\mathbf{c}$). } \label{fig:treelstm}
\end{figure}

After the AR-Tree construction, we use Tree-LSTM~\cite{tai2015improved,zhu2015long}, which introduces \emph{cell state} into TreeRNNs to achieve better information flow, as the composition function to compute parent representation from its children and corresponding word in a bottom-up manner (\textit{i.e.}, Figure~\ref{fig:treelstm}).
Because the original word sequence is kept in the in-order traversal of the AR-Tree, Tree-LSTM units can utilize both the sequential and the structural information to compose semantics.

The complete Tree-LSTM composition function in our model is as follows:

\begin{equation}
\begin{array}{c}
\begin{bmatrix}\mathbf{i} \\ \mathbf{f_L} \\ \mathbf{f_R} \\ \mathbf{f_i} \\ \mathbf{o} \\ \mathbf{g} \end{bmatrix} = 
\begin{bmatrix} \sigma \\ \sigma \\ \sigma \\ \sigma \\ \sigma \\ tanh \end{bmatrix}
\Big(\mathbf{W_c} \begin{bmatrix} \mathbf{h_L} \\ \mathbf{h_R} \\ \mathbf{h_i} \end{bmatrix} + \mathbf{b_c}\Big), \\
\mathbf{c}=\mathbf{f_L} \odot \mathbf{c_L} + \mathbf{f_R} \odot \mathbf{c_R} + \mathbf{f_i} \odot \mathbf{c_i} + \mathbf{i} \odot \mathbf{g}, \\
\mathbf{h}=\mathbf{o} \odot tanh(\mathbf{c}), \\
\end{array}
\end{equation}
where $(\mathbf{h_L}$,$\mathbf{c_L})$, $(\mathbf{h_R}$,$\mathbf{c_R})$ and $(\mathbf{h_i}$,$\mathbf{c_i})$ come from left child, right child, and bidirectional LSTM, respectively. For those nodes missing some inputs, such as the leaf nodes or nodes with only one child, we fill the missing inputs with zeros.

Finally, we use $\mathbf{h}$ of the root $R$ as the embedding of the sentence $S$ and feed it into downstream tasks. The sentence embedding will focus on those informative words as they are closer to root and their semantics is emphasized naturally.

\section{End-to-end Training Using REINFORCE}\label{sec:rl}
Our overall training loss $\mathcal{L}$ combines the loss of the downstream task $\mathcal{L}_{task}$ (\textit{e.g.}, the cross-entropy loss for classification tasks), the tree construction loss $\mathcal{L}_{tree}=-J(\theta)$ (discussed soon), and an L2 regularization term on all trainable parameters $\phi$:
\begin{equation}\label{loss}
\mathcal{L} = \mathcal{L}_{task} + \alpha \mathcal{L}_{tree} + \lambda ||\phi||_2^2,
\end{equation}
where $\alpha$ and $\lambda$ are trade-off hyperparameters.

We train the model only according to the downstream task and do not incorporate any structure supervision or pre-trained parser, leading to non-differentiability as the AR-Tree construction is a discrete decision-making process. 
Specifically, the scoring function $Score(\mathbf{h})$ cannot be learned in an end-to-end manner when optimizing $\mathcal{L}_{task}$.
Inspired by \cite{yogatama2016learning}, we employ reinforcement learning, whose objective corresponds to $\mathcal{L}_{tree}$, to train the scoring function. 

We consider the construction of AR-Tree as a recursive decision-making process where each action selects a word for a node. For node $t \in T$, we define the \textbf{\emph{state}} $s_t$ as its corresponding sequence $S[b_t:e_t]$, where $b_t$ and $e_t$ respectively represent the index of beginning and ending position. 
The \textbf{\emph{action space}} $A_t$ is $\{b_t, b_t+1, \cdots, e_t\}$. 
We feed scores of candidate words into a softmax layer as our \textbf{\emph{policy network}}, which outputs a probability distribution over the action space:
\begin{equation}
\pi(a_t=i|s_t; \theta) = \frac{\text{exp}(Score(\mathbf{h}_i; \theta))}{\sum_{j=b_t}^{e_t} \text{exp}(Score(\mathbf{h}_j; \theta))},
\end{equation}
where $i \in A_t$. Different from Algorithm~\ref{construct} which is greedy and deterministic, at training, we construct AR-Trees randomly in terms of $\pi$, to explore more structures and bring greater gain in the long run. 
After the \textbf{\emph{action}} $a_t$ is sampled based on $\pi(a_t|s_t; \theta)$, the sequence is split into $S[b_t:a_t-1]$ and $S[a_t+1:e_t]$, which are used respectively as two children's states.

As for the \textbf{\emph{reward}} $r_t$, we consider the performance metric on a downstream task. 
For simplicity, we discuss a classification task in this paper and leave further explorations in future work.
After the whole tree $T$ is recursively sampled based on $\pi$, we can obtain the sentence embedding following Section~\ref{sec:bottom-up}, feed it into the downstream classifier and obtain a predicted label.
The sampled tree is considered good if the prediction is correct, and bad if the prediction is wrong.
A simple rewarding strategy is to give $r_t=1$ for all $t$ in a good tree, and $r_t=-1$ for all $t$ in a bad tree.

However, we consider that an early decision from a longer sequence has a greater impact on the tree structure (\textit{e.g.}, the selection of root is more important than leaves).
So we multiply the reward value by $|A_t|$, \textit{i.e.}, $r_t=|A_t|$ for all $t$ in a good tree and $r_t=-|A_t|$ for all $t$ in a bad tree.

We use REINFORCE~\cite{williams1992simple}, a widely-used policy gradient method in reinforcement learning, to learn parameters of the policy network. The goal of the learning algorithm is to maximize the expected long-term reward:
\begin{equation}
J(\theta) = \mathbb{E}_{\pi(a_t|s_t; \theta)} r_t.
\end{equation}
Following \cite{sutton2000policy}, the gradient w.r.t. the parameters of policy network can be derived as:
\begin{equation}
\nabla_{\theta} J(\theta) = \mathbb{E}_{\pi} [r_t \nabla_{\theta} \log \pi(a_t|s_t; \theta)].
\end{equation}

It is prohibitively expensive to calculate the accurate $\nabla_{\theta} J(\theta)$ by iterating over all possible trees. 
Following \cite{yu2017seqgan}, we apply Monte Carlo search to estimate the expectation. 
Specifically, we sample $M$ trees for sentence $S$, denoted as $T_1, \cdots, T_M$, each containing $N$ nodes. 
Then we can simplify $\nabla_{\theta} J(\theta)$ by averaging rewards among all these $M \times N$ nodes (micro average):
\begin{equation}\label{formula:node-level}
\nabla_{\theta} J(\theta) = \frac{1}{MN} \sum_{k=1}^M \sum_{t \in T_k} r_t \nabla_{\theta} \log \pi(a_t|s_t;\theta).
\end{equation}

However, we observed that frequent words (\textit{e.g.}, \emph{the, is}) were assigned high scores if we used Formula~\ref{formula:node-level} to train $\theta$. 
We think the reason is that the scoring function takes one single word embedding as input, meaning that frequent words will contribute more to its training gradient if rewards are averaged among all tree nodes, which is harmful to the score estimation of low-frequency words. 

\begin{figure}[h]
\includegraphics[width=\linewidth]{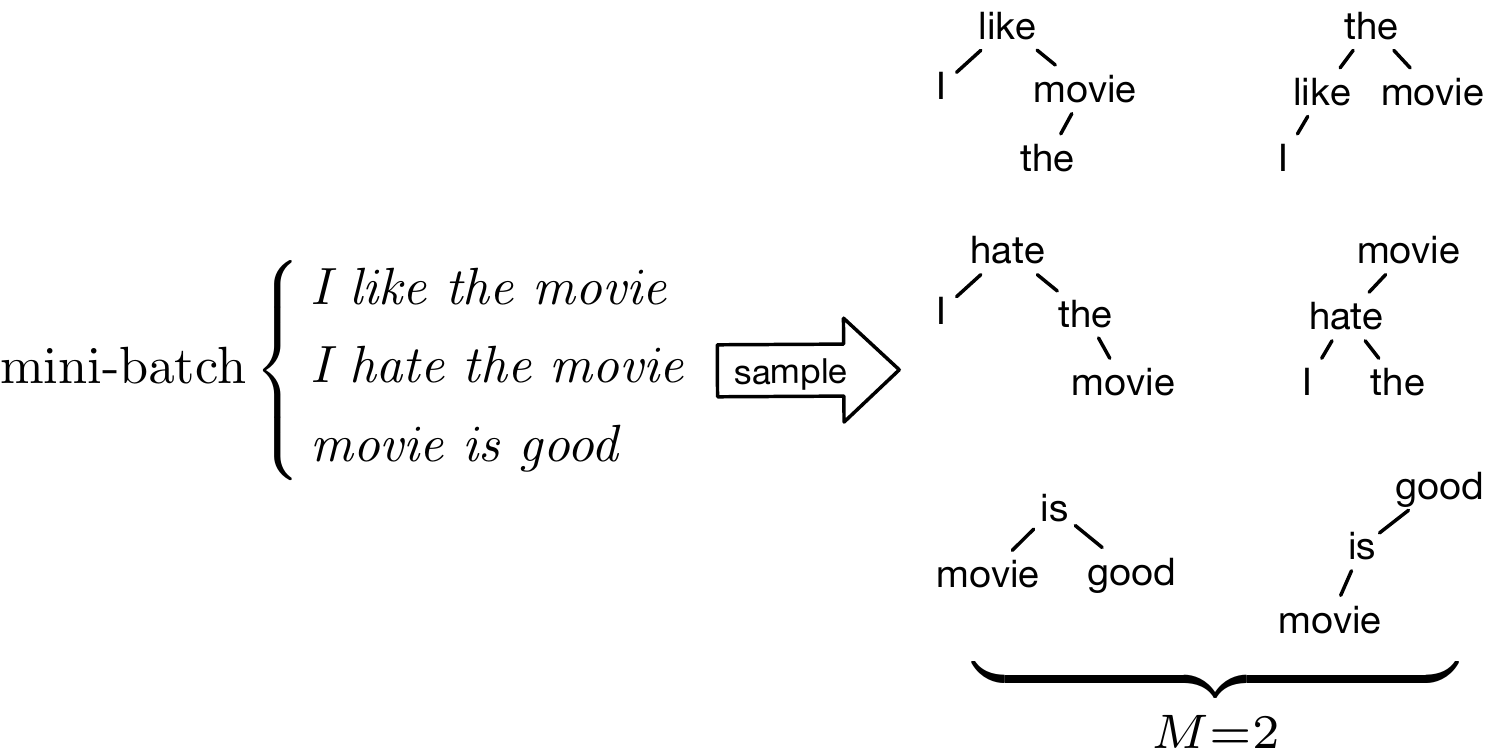}
\caption{Sampled results of a mini-batch. We have three sentences in a mini-batch ($B=3$) and for each sentence we sample two trees ($M=2$). Totally we get six sampled trees. Micro average is to average gradients over all nodes of these six trees. Macro average is to first average gradients over nodes of the same word (e.g., average over 6 nodes containing \emph{movie} to get the gradient of \emph{movie}), and then average gradients of these words to obtain the final training signals. \label{fig:batch}}
\end{figure}

To eliminate the influence caused by the word frequency, we integrate $B$ input sentences as a mini-batch, sample $M$ trees for each of them, and normalize the gradient in word-level (macro average) rather than node-level:

\begin{equation}\label{formula:word-level}
\nabla_{\theta}\! J(\theta) \!=\! \frac{1}{|W|}\!\sum_{w \in W} \frac{1}{|T^w|}\!\sum_{t \in T^w}\!\! r_t \nabla_{\theta} \! \log \pi(a_t|s_t;\theta),
\end{equation}
where $W$ represents all words of the mini-batch, $T^w$ represents all nodes whose selected word is $w$ in all $B \times M$ sampled trees. Figure~\ref{fig:batch} gives an example.

\section{Experiments}

\begin{table*}[!ht]
\small
\centering
\begin{tabular}{| l | c | c | c | c | c | c | c | c| l |}
\hline
{Experiment} & {$D_x$} & {$D_h$} & {$D_c$} & {Finetune} & {Dropout} & {Bn} & {Batch size} & {$M$} & {Optimizer}\\
\hline
SNLI-100D      & 100 & 100 & 200 & $\surd$ & 0.1 & $\surd$ & 128 & 2 & Adam~\cite{kingma2014adam}\\
SNLI-300D      & 300 & 300 & 1024&         & 0.1 & $\surd$ & 128 & 2 & Adam\\
SST-2          & 300 & 300 & 300 & $\surd$ & 0.5 &         & 32  & 3 & Adadelta~\cite{zeiler2012adadelta}\\
SST-5          & 300 & 300 & 1024& $\surd$ & 0.5 &         & 64  & 3 & Adadelta\\
Age prediction & 300 & 600 & 2000& $\surd$ & 0.3 &         & 50  & 3 & Adam\\
\hline
\end{tabular}
\caption{\label{setting} Experimental settings. Dropout: dropout probability. Bn: whether using batch normalization. } 
\end{table*}

\begin{table*}[ht]
\small
\centering
\begin{tabular}{|l|r|r|}
\hline
{Model} & {\# params.} & {Acc. (\%)} \\
\hline
\hline
100D Latent Syntax Tree-LSTM~\cite{yogatama2016learning} & 500k & 80.5 \\
100D CYK Tree-LSTM~\cite{maillard2017jointly} & 231k & 81.6 \\
100D Gumbel Tree-LSTM~\cite{choi2017learning}  & 262k & 82.6 \\
100D Tf-idf Tree-LSTM (Ours) & 343k & 82.3 \\
100D AR-Tree (Ours) & 356k & \textbf{82.8} \\
\hline
\hline
300D SPINN~\cite{bowman2016fast}  & 3.7m & 83.2 \\
300D NSE~\cite{munkhdalai2017nse}  & 6.3m & 84.8 \\
300D NTI-SLSTM-LSTM~\cite{munkhdalai2017neural} & 4.0m & 83.4 \\
300D Gumbel Tree-LSTM~\cite{choi2017learning} & 2.9m & 85.0 \\
300D Self-Attentive~\cite{lin2017structured} & 4.1m & 84.4 \\
300D Tf-idf Tree-LSTM (Ours) & 3.5m & 84.5 \\
300D AR-Tree (Ours) & 3.6m & \textbf{85.5} \\
\hline
\hline
600D Gated-Attention BiLSTM~\cite{chen2017recurrent} & 11.6m & 85.5 \\
300D Decomposable attention~\cite{parikh2016decomposable} & 582k & 86.8 \\
300D NTI-SLSTM-LSTM global attention~\cite{munkhdalai2017neural} & 3.2m & \textbf{87.3} \\
300D Structured Attention~\cite{kim2017structured} & 2.4m & 86.8 \\
\hline
\end{tabular}
\caption{\label{snli} Test accuracy and the number of parameters (excluding word embeddings) on the SNLI dataset. The above two sections list results of Tree-LSTM and other baseline models grouped by the dimension. The bottom section contains state-of-the-art inter-attention models on SNLI dataset.}
\end{table*}

We evaluate the proposed AR-Tree on three tasks: natural language inference, sentence sentiment analysis, and author profiling.

We set $\alpha = 0.1$, $\lambda = 1e-5$ in Eq.~\ref{loss} through all experiments.
For fair comparisons, we followed the experimental settings in \cite{choi2017learning} on language inference and sentence sentiment analysis. 
For the author profiling task whose dataset is provided by~\cite{lin2017structured}, we followed their settings by contacting the authors.
We considered their model, which is self-attentive but without tree structures, as a baseline, to show the effect of latent trees.
We conducted \textit{Tf-idf Tree-LSTM} experiment, which replaces the scoring function with tf-idf value while retaining all other settings, as one of our baselines. 
For all experiments, we saved the model that performed best on the validation set as our final model and evaluated it on the test set.
The implementation is made publicly available.\footnote{\url{https://github.com/shijx12/AR-Tree}}

\subsection{Natural Language Inference}

The natural language inference is a task of predicting the semantic relationship between two sentences, a premise, and a hypothesis. We evaluated our model using the Stanford Natural Language Inference corpus (SNLI; \cite{bowman2015large}), which aims to predict whether two sentences are \emph{entailment}, \emph{contradiction}, or \emph{neutral}. SNLI consists of 549,367/9,842/9,824 premise-hypothesis pairs for train/validation/test sets respectively.

Following \cite{bowman2016fast,mou2016natural}, we ran AR-Tree separately on two input sentences to obtain their embeddings $\mathbf{h}_{pre}$ and $\mathbf{h}_{hyp}$. Then we constructed a feature vector $\mathbf{v}$ for the pair by the following equation:
\begin{equation}
\mathbf{v} = \begin{bmatrix}\mathbf{h}_{pre} \\ \mathbf{h}_{hyp} \\ |\mathbf{h}_{pre} - \mathbf{h}_{hyp}| \\ \mathbf{h}_{pre} \odot \mathbf{h}_{hyp} \end{bmatrix},
\end{equation}
 and fed the feature into a neural network, \textit{i.e.}, a multi-layer perceptron (MLP) which has a $D_c$-dimentional hidden layer with \emph{ReLU} activation function and a \emph{softmax} layer.

We conducted SNLI experiments with two settings: 100D ($D_x=100$) and 300D ($D_x=300$). In both experiments, we initialized the word embedding matrix with GloVe pre-trained vectors~\cite{pennington2014glove}, added a dropout~\cite{srivastava2014dropout} after the word embedding layer and added batch normalization layers~\cite{ioffe2015batch} followed by dropout to the input and the output of the MLP. Details can be found in Table~\ref{setting}. The training on an NVIDIA GTX1080 Ti needs about 30 hours, slower than Gumbel Tree-LSTM~\cite{choi2017learning} because our tree construction is implemented for every single sentence instead of the whole batch.

\begin{table*}[!ht]
\small
\centering
\begin{tabular}{|l|r|r|}
\hline
{Model} & {SST-2 (\%)} & {SST-5 (\%)} \\
\hline
\hline
LSTM~\cite{tai2015improved} & 84.9 & 46.4 \\
Bidirectional LSTM~\cite{tai2015improved} & 87.5 & 49.1 \\
RNTN~\cite{socher2013recursive} & 85.4 & 45.7 \\
DMN~\cite{kumar2016ask} & 88.6 & 52.1 \\
NSE~\cite{munkhdalai2017nse} & 89.7 & 52.8 \\
BCN+Char+CoVe~\cite{mccann2017learned} & 90.3 & \textbf{53.7} \\
byte-mLSTM~\cite{radford2017learning} & \textbf{91.8} & 52.9 \\
\hline
\hline
Constituency Tree-LSTM~\cite{tai2015improved} & 88.0 & 51.0 \\
Latent Syntax Tree-LSTM~\cite{yogatama2016learning} & 86.5 & - \\
NTI-SLSTM-LSTM~\cite{munkhdalai2017neural} & 89.3 & \textbf{53.1} \\
Gumble Tree-LSTM~\cite{choi2017learning} & 90.1 & 52.5 \\
Tf-idf Tree-LSTM (Ours) & 88.9 & 51.3 \\
AR-Tree (Ours) & \textbf{90.4} & 52.7 \\
\hline
\end{tabular}
\caption{\label{sst} Results of SST experiments. The bottom section contains results of Tree-LSTM models and the top section contains other baseline and state-of-the-art models.}
\end{table*}

Table~\ref{snli} summarizes the results. We can see that our 100D and 300D models perform best among the Tree-LSTM models. State-of-the-art inter-attention models get the highest performance on SNLI, because they incorporate inter-information between the sentence pairs to boost the performance. However, inter-attention is limited for paired inputs and lacks flexibility. Our 300D model outperforms self-attentive~\cite{lin2017structured}, the state-of-the-art intra-attention model, by 1.1\%, demonstrating its effectiveness.

\subsection{Sentiment Analysis}
We used Stanford Sentiment Treebank (SST)~\cite{socher2013recursive} to evaluate the performance of our model. The sentences in SST dataset are parsed into binary trees with the Stanford parser, and each subtree corresponding to a phrase is annotated with a sentiment score. It includes two subtasks: SST-5, classifying each phrase into 5 classes, and SST-2, preserving only 2 classes.

Following \cite{choi2017learning}, we used all phrases for training but only the entire sentences for evaluation. We used an MLP with a $D_c$-dimensional hidden layer as the classifier.  For both SST-2 and SST-5, we initialized the word embeddings with GloVe 300D pre-trained vectors, and added dropout to the word embedding layer and the input and the output of the MLP. Table~\ref{setting} lists the parameter details.

Table~\ref{sst} shows the results of SST experiments. Our model on SST-2 outperforms all Tree-LSTM models and other state-of-the-art models except Byte-mLSTM~\cite{radford2017learning}, a byte-level language model trained on a very large corpus. \cite{mccann2017learned} obtains the highest performance on SST-5 due to the help of pretraining and character n-gram embeddings. Without the help of character-level information, our model can still get comparable results on SST-5.

\subsection{Author Profiling}
The Author Profiling dataset consists of Twitter tweets and some annotations about age and gender of the user writing the tweet. Following \cite{lin2017structured} we used English tweets as input to predict the age range of the user, including 5 classes: 18-24, 25-34, 35-49, 50-64 and 65+. The \emph{age prediction} dataset consists of 68,485/4,000/4,000 tweets for train/validation/test sets. 

We applied GloVe and dropout as in the SST experiments. Table~\ref{setting} describes detailed settings, which are the same as \cite{lin2017structured}'s published implementation except for the optimizer (they use SGD but we find Adam converges better).

\begin{table}[!ht]
\footnotesize
\centering
\begin{tabular}{|l|r|}
\hline
{Model} & {Acc. (\%)} \\
\hline
BiLSTM+MaxPooling~\cite{lin2017structured} & 77.40 \\
CNN+MaxPooling~\cite{lin2017structured} & 78.15 \\
{\fontsize{8.5pt}{8.5pt}\selectfont Gumble Tree-LSTM~\cite{choi2017learning}} & 80.23 \\
Self-Attentive~\cite{lin2017structured} & 80.45 \\
Tf-idf Tree-LSTM (Ours) & 80.20 \\
AR-Tree (Ours) & \textbf{80.85} \\
\hline
\end{tabular}
\caption{\label{age} Results of age prediction experiments. }
\end{table}

Results of the age prediction experiments are shown in Table~\ref{age}. We can see that our model outperforms all other baseline models. Compared to self-attentive model, our AR-Tree model obtains higher performance in the same experimental settings, indicating that latent structures are helpful to sentence understanding.

\begin{figure*}
\includegraphics[width=\textwidth]{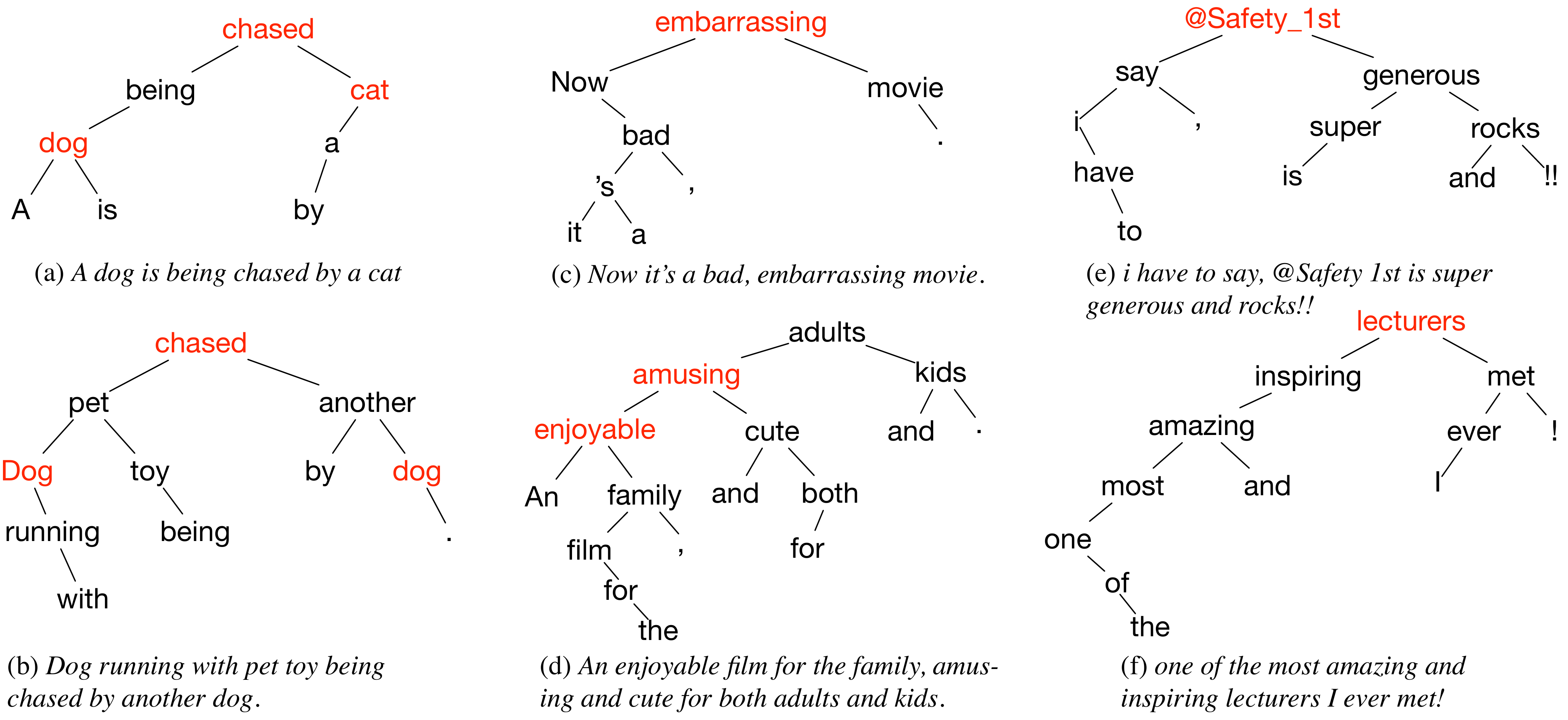}
\caption{\label{fig:quality}Examples of our produced attentive trees. The caption of each subfigure is the input sentence. The left, middle and right columns are from SNLI, SST-2 and age prediction respectively. We can see that our AR-Tree can place task-informative words at shallow nodes.}
\end{figure*}

\begin{figure*}[ht]
\includegraphics[width=\linewidth]{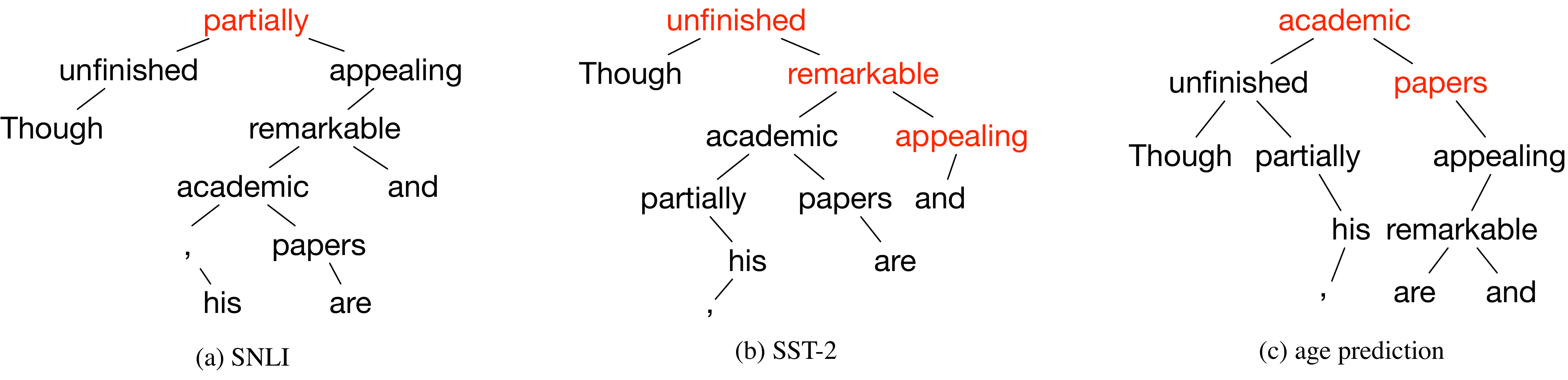}
\caption{Different structures from different trained parsers for the same sentence \textit{Though unfinished partially, his academic papers are remarkable and appealing}. We can see that words are emphasized adaptively based on the target task.}
\label{fig:diff}
\end{figure*}


\section{Qualitative Analysis}

We conducted experiments to observe structures of the learned trees.
We select 2 sentences from the test set of three experiment datasets respectively and show their attentive trees in Figure~\ref{fig:quality}.

The left column is a sentence pair with relationship \emph{contradiction} from SNLI. Figure~\ref{fig:quality}a and \ref{fig:quality}b both focus on the predicate word \emph{chased} firstly, then focus on its subject and object respectively.
The middle column is from SST-2, the sentiment analysis dataset. Both Figure~\ref{fig:quality}c and \ref{fig:quality}d focus on emotional adjectives such as \emph{embarrassing}, \emph{amusing} and \emph{enjoyable}.
The right column is from the age prediction dataset, predicting the author's age based on the tweet. Figure \ref{fig:quality}e attends to \emph{@Safety\_1st}, a baby production, indicating that the author is probably a young parent. Figure \ref{fig:quality}f focuses on \emph{lecturers} which suggests that the author is likely to be a college student.

Furthermore, we applied parsers trained on different tasks to the same sentence, and show results in Figure~\ref{fig:diff}. The parser of SNLI focuses on \emph{partially} (Figure~\ref{fig:diff}a), as SNLI is an inference dataset and pays more attention to words which may be different in two sentences to reflect the contradiction relationship (\textit{e.g.}, \emph{partially} v.s. \emph{totally}). The parser of SST-2, the sentiment classification task, focuses on sentimental words (Figure~\ref{fig:diff}b) as we have expected. In the parsed results of age prediction, \emph{academic} and \emph{papers} are emphasized (Figure~\ref{fig:diff}c) because they are more likely to be discussed by college students, and are more informative to the age prediction task than other words.

Our model is able to pay attention to task-specific critical words for different tasks and learn interpretable structures, which is beneficial to the sentence understanding.


\section{Conclusions and Future Work}
We propose Attentive Recursive Tree (AR-Tree), a novel yet generic latent Tree-LSTM sentence embedding model, learning to learn task-specific structural embedding guided by word attention. 
Results on three different datasets demonstrate that AR-Tree learns reasonable attentive tree structures and outperforms previous Tree-LSTM models.

Moving forward, we are going to design a batch-mode tree construction algorithm, \textit{e.g.}, asynchronous parallel recursive tree construction, to make the full exploitation of distributed and parallel computing power. 
Therefore, we may able to learn an AR-Forest to embed paragraphs.

\section{Acknowledgments}
The work is supported by National Key Research and Development Program of China (2017YFB1002101), NSFC key project (U1736204, 61533018), and THUNUS NExT Co-Lab.

\fontsize{9.5pt}{10.5pt} \selectfont
\bibliography{reference_674}

\begin{thebibliography}{}

\bibitem[\protect\citeauthoryear{Arora, Liang, and Ma}{2016}]{arora2016simple}
Arora, S.; Liang, Y.; and Ma, T.
\newblock 2016.
\newblock A simple but tough-to-beat baseline for sentence embeddings.

\bibitem[\protect\citeauthoryear{Blunsom, Grefenstette, and
  Kalchbrenner}{2014}]{blunsom2014convolutional}
Blunsom, P.; Grefenstette, E.; and Kalchbrenner, N.
\newblock 2014.
\newblock A convolutional neural network for modelling sentences.
\newblock In {\em ACL}.

\bibitem[\protect\citeauthoryear{Bowman \bgroup et al\mbox.\egroup
  }{2015}]{bowman2015large}
Bowman, S.~R.; Angeli, G.; Potts, C.; and Manning, C.~D.
\newblock 2015.
\newblock A large annotated corpus for learning natural language inference.
\newblock In {\em EMNLP}.

\bibitem[\protect\citeauthoryear{Bowman \bgroup et al\mbox.\egroup
  }{2016}]{bowman2016fast}
Bowman, S.~R.; Gauthier, J.; Rastogi, A.; Gupta, R.; Manning, C.~D.; and Potts,
  C.
\newblock 2016.
\newblock A fast unified model for parsing and sentence understanding.
\newblock In {\em ACL}.

\bibitem[\protect\citeauthoryear{Chen \bgroup et al\mbox.\egroup
  }{2017}]{chen2017recurrent}
Chen, Q.; Zhu, X.; Ling, Z.-H.; Wei, S.; Jiang, H.; and Inkpen, D.
\newblock 2017.
\newblock Recurrent neural network-based sentence encoder with gated attention
  for natural language inference.
\newblock In {\em RepEval}.

\bibitem[\protect\citeauthoryear{Choi, Yoo, and goo
  Lee}{2017}]{choi2017learning}
Choi, J.; Yoo, K.~M.; and goo Lee, S.
\newblock 2017.
\newblock Learning to compose task-specific tree structures.
\newblock In {\em AAAI}.

\bibitem[\protect\citeauthoryear{Cocke}{1970}]{cocke1970programming}
Cocke, J.
\newblock 1970.
\newblock Programming languages and their compilers: Preliminary notes.

\bibitem[\protect\citeauthoryear{Dai and Le}{2015}]{dai2015semi}
Dai, A.~M., and Le, Q.~V.
\newblock 2015.
\newblock Semi-supervised sequence learning.
\newblock In {\em NIPS}.

\bibitem[\protect\citeauthoryear{dos Santos \bgroup et al\mbox.\egroup
  }{2016}]{dos2016attentive}
dos Santos, C.~N.; Tan, M.; Xiang, B.; and Zhou, B.
\newblock 2016.
\newblock Attentive pooling networks.
\newblock {\em CoRR, abs/1602.03609}.

\bibitem[\protect\citeauthoryear{Goldberg and
  Elhadad}{2010}]{goldberg2010efficient}
Goldberg, Y., and Elhadad, M.
\newblock 2010.
\newblock An efficient algorithm for easy-first non-directional dependency
  parsing.
\newblock In {\em NAACL}.

\bibitem[\protect\citeauthoryear{Hill, Cho, and
  Korhonen}{2016}]{hill2016learning}
Hill, F.; Cho, K.; and Korhonen, A.
\newblock 2016.
\newblock Learning distributed representations of sentences from unlabelled
  data.
\newblock In {\em NAACL}.

\bibitem[\protect\citeauthoryear{Hu \bgroup et al\mbox.\egroup
  }{2014}]{hu2014convolutional}
Hu, B.; Lu, Z.; Li, H.; and Chen, Q.
\newblock 2014.
\newblock Convolutional neural network architectures for matching natural
  language sentences.
\newblock In {\em NIPS}.

\bibitem[\protect\citeauthoryear{Ioffe and Szegedy}{2015}]{ioffe2015batch}
Ioffe, S., and Szegedy, C.
\newblock 2015.
\newblock Batch normalization: Accelerating deep network training by reducing
  internal covariate shift.
\newblock In {\em ICML}.

\bibitem[\protect\citeauthoryear{Jang, Gu, and
  Poole}{2016}]{jang2016categorical}
Jang, E.; Gu, S.; and Poole, B.
\newblock 2016.
\newblock Categorical reparameterization with gumbel-softmax.
\newblock {\em arXiv preprint arXiv:1611.01144}.

\bibitem[\protect\citeauthoryear{Kasami}{1965}]{kasami1965efficient}
Kasami, T.
\newblock 1965.
\newblock An efficient recognition and syntaxanalysis algorithm for
  context-free languages.
\newblock Technical report.

\bibitem[\protect\citeauthoryear{Kim \bgroup et al\mbox.\egroup
  }{2017}]{kim2017structured}
Kim, Y.; Denton, C.; Hoang, L.; and Rush, A.~M.
\newblock 2017.
\newblock Structured attention networks.
\newblock {\em arXiv preprint arXiv:1702.00887}.

\bibitem[\protect\citeauthoryear{Kingma and Ba}{2014}]{kingma2014adam}
Kingma, D., and Ba, J.
\newblock 2014.
\newblock Adam: A method for stochastic optimization.
\newblock {\em arXiv preprint arXiv:1412.6980}.

\bibitem[\protect\citeauthoryear{Kumar \bgroup et al\mbox.\egroup
  }{2016}]{kumar2016ask}
Kumar, A.; Irsoy, O.; Ondruska, P.; Iyyer, M.; Bradbury, J.; Gulrajani, I.;
  Zhong, V.; Paulus, R.; and Socher, R.
\newblock 2016.
\newblock Ask me anything: Dynamic memory networks for natural language
  processing.
\newblock In {\em ICML}.

\bibitem[\protect\citeauthoryear{Lin \bgroup et al\mbox.\egroup
  }{2017}]{lin2017structured}
Lin, Z.; Feng, M.; Santos, C. N.~d.; Yu, M.; Xiang, B.; Zhou, B.; and Bengio,
  Y.
\newblock 2017.
\newblock A structured self-attentive sentence embedding.
\newblock {\em arXiv preprint arXiv:1703.03130}.

\bibitem[\protect\citeauthoryear{Maillard, Clark, and
  Yogatama}{2017}]{maillard2017jointly}
Maillard, J.; Clark, S.; and Yogatama, D.
\newblock 2017.
\newblock Jointly learning sentence embeddings and syntax with unsupervised
  tree-lstms.
\newblock {\em arXiv preprint arXiv:1705.09189}.

\bibitem[\protect\citeauthoryear{McCann \bgroup et al\mbox.\egroup
  }{2017}]{mccann2017learned}
McCann, B.; Bradbury, J.; Xiong, C.; and Socher, R.
\newblock 2017.
\newblock Learned in translation: Contextualized word vectors.
\newblock In {\em NIPS}.

\bibitem[\protect\citeauthoryear{Mikolov \bgroup et al\mbox.\egroup
  }{2013}]{mikolov2013distributed}
Mikolov, T.; Sutskever, I.; Chen, K.; Corrado, G.~S.; and Dean, J.
\newblock 2013.
\newblock Distributed representations of words and phrases and their
  compositionality.
\newblock In {\em NIPS}.

\bibitem[\protect\citeauthoryear{Mou \bgroup et al\mbox.\egroup
  }{2016}]{mou2016natural}
Mou, L.; Men, R.; Li, G.; Xu, Y.; Zhang, L.; Yan, R.; and Jin, Z.
\newblock 2016.
\newblock Natural language inference by tree-based convolution and heuristic
  matching.
\newblock In {\em ACL}.

\bibitem[\protect\citeauthoryear{Munkhdalai and Yu}{2017a}]{munkhdalai2017nse}
Munkhdalai, T., and Yu, H.
\newblock 2017a.
\newblock Neural semantic encoders.
\newblock In {\em ACL}.

\bibitem[\protect\citeauthoryear{Munkhdalai and
  Yu}{2017b}]{munkhdalai2017neural}
Munkhdalai, T., and Yu, H.
\newblock 2017b.
\newblock Neural tree indexers for text understanding.
\newblock In {\em ACL}.

\bibitem[\protect\citeauthoryear{Nivre}{2003}]{nivre2003dependency}
Nivre, J.
\newblock 2003.
\newblock An efficient algorithm for projective dependency parsing.
\newblock In {\em Proceedings of the 8th International Workshop on Parsing
  Technologies (IWPT}.

\bibitem[\protect\citeauthoryear{Parikh \bgroup et al\mbox.\egroup
  }{2016}]{parikh2016decomposable}
Parikh, A.; T{\"a}ckstr{\"o}m, O.; Das, D.; and Uszkoreit, J.
\newblock 2016.
\newblock A decomposable attention model for natural language inference.
\newblock In {\em EMNLP}.

\bibitem[\protect\citeauthoryear{Pennington, Socher, and
  Manning}{2014}]{pennington2014glove}
Pennington, J.; Socher, R.; and Manning, C.
\newblock 2014.
\newblock Glove: Global vectors for word representation.
\newblock In {\em EMNLP}.

\bibitem[\protect\citeauthoryear{Radford, Jozefowicz, and
  Sutskever}{2017}]{radford2017learning}
Radford, A.; Jozefowicz, R.; and Sutskever, I.
\newblock 2017.
\newblock Learning to generate reviews and discovering sentiment.
\newblock {\em arXiv preprint arXiv:1704.01444}.

\bibitem[\protect\citeauthoryear{Shi \bgroup et al\mbox.\egroup
  }{2018}]{shi2018tree}
Shi, H.; Zhou, H.; Chen, J.; and Li, L.
\newblock 2018.
\newblock On tree-based neural sentence modeling.
\newblock In {\em Proceedings of the 2018 Conference on Empirical Methods in
  Natural Language Processing},  4631--4641.

\bibitem[\protect\citeauthoryear{Socher \bgroup et al\mbox.\egroup
  }{2011}]{socher2011semi}
Socher, R.; Pennington, J.; Huang, E.~H.; Ng, A.~Y.; and Manning, C.~D.
\newblock 2011.
\newblock Semi-supervised recursive autoencoders for predicting sentiment
  distributions.
\newblock In {\em EMNLP}.

\bibitem[\protect\citeauthoryear{Socher \bgroup et al\mbox.\egroup
  }{2013}]{socher2013recursive}
Socher, R.; Perelygin, A.; Wu, J.; Chuang, J.; Manning, C.~D.; Ng, A.; and
  Potts, C.
\newblock 2013.
\newblock Recursive deep models for semantic compositionality over a sentiment
  treebank.
\newblock In {\em EMNLP}.

\bibitem[\protect\citeauthoryear{Srivastava \bgroup et al\mbox.\egroup
  }{2014}]{srivastava2014dropout}
Srivastava, N.; Hinton, G.~E.; Krizhevsky, A.; Sutskever, I.; and
  Salakhutdinov, R.
\newblock 2014.
\newblock Dropout: a simple way to prevent neural networks from overfitting.
\newblock {\em JMLR}.

\bibitem[\protect\citeauthoryear{Sutton \bgroup et al\mbox.\egroup
  }{2000}]{sutton2000policy}
Sutton, R.~S.; McAllester, D.~A.; Singh, S.~P.; and Mansour, Y.
\newblock 2000.
\newblock Policy gradient methods for reinforcement learning with function
  approximation.
\newblock In {\em NIPS}.

\bibitem[\protect\citeauthoryear{Tai, Socher, and
  Manning}{2015}]{tai2015improved}
Tai, K.~S.; Socher, R.; and Manning, C.~D.
\newblock 2015.
\newblock Improved semantic representations from tree-structured long
  short-term memory networks.
\newblock In {\em ACL}.

\bibitem[\protect\citeauthoryear{Wang and Nyberg}{2015}]{wang2015long}
Wang, D., and Nyberg, E.
\newblock 2015.
\newblock A long short-term memory model for answer sentence selection in
  question answering.
\newblock In {\em ACL}.

\bibitem[\protect\citeauthoryear{Williams, Drozdov, and
  Bowman}{2017}]{williams2017learning}
Williams, A.; Drozdov, A.; and Bowman, S.~R.
\newblock 2017.
\newblock Learning to parse from a semantic objective: It works. is it syntax?
\newblock {\em arXiv preprint arXiv:1709.01121}.

\bibitem[\protect\citeauthoryear{Williams}{1992}]{williams1992simple}
Williams, R.~J.
\newblock 1992.
\newblock Simple statistical gradient-following algorithms for connectionist
  reinforcement learning.
\newblock {\em Machine learning}.

\bibitem[\protect\citeauthoryear{Yogatama \bgroup et al\mbox.\egroup
  }{2016}]{yogatama2016learning}
Yogatama, D.; Blunsom, P.; Dyer, C.; Grefenstette, E.; and Ling, W.
\newblock 2016.
\newblock Learning to compose words into sentences with reinforcement learning.
\newblock {\em arXiv preprint arXiv:1611.09100}.

\bibitem[\protect\citeauthoryear{Younger}{1967}]{younger1967recognition}
Younger, D.~H.
\newblock 1967.
\newblock Recognition and parsing of context-free languages in time n3.
\newblock {\em Information and control}.

\bibitem[\protect\citeauthoryear{Yu \bgroup et al\mbox.\egroup
  }{2017}]{yu2017seqgan}
Yu, L.; Zhang, W.; Wang, J.; and Yu, Y.
\newblock 2017.
\newblock Seqgan: Sequence generative adversarial nets with policy gradient.
\newblock In {\em AAAI}.

\bibitem[\protect\citeauthoryear{Zeiler}{2012}]{zeiler2012adadelta}
Zeiler, M.~D.
\newblock 2012.
\newblock Adadelta: an adaptive learning rate method.
\newblock {\em arXiv preprint arXiv:1212.5701}.

\bibitem[\protect\citeauthoryear{Zhu, Sobihani, and Guo}{2015}]{zhu2015long}
Zhu, X.; Sobihani, P.; and Guo, H.
\newblock 2015.
\newblock Long short-term memory over recursive structures.
\newblock In {\em ICML}.

\end{thebibliography}
\bibliographystyle{aaai}
\end{document}